\documentclass[11pt,a4paper]{article}
\usepackage[T1]{fontenc}
\usepackage[utf8]{inputenc}
\usepackage{newtxtext,newtxmath}
\usepackage[a4paper,top=25mm,bottom=25mm,left=30mm,right=30mm]{geometry}
\usepackage{microtype}
\usepackage{graphicx}
\usepackage{xcolor}
\usepackage{hyperref}
\usepackage{caption}
\usepackage{fancyhdr}
\graphicspath{{BlockPython_Current_System_Report_assets/}}
\definecolor{BlockBlue}{HTML}{264E86}
\hypersetup{colorlinks=true,linkcolor=BlockBlue,urlcolor=BlockBlue,citecolor=BlockBlue,pdftitle={BlockPython Current System Report}}
\title{\textbf{BlockPython: A Process-Aware Agent-Supported Platform for the Transition from Block-Based to Python Programming}}
\author{Zexi Chen$^{1}$, Haoming Wang$^{2,1}$, Mingwei Xu$^{3}$, Xianlong Xu$^{1}$\\[0.5em]
\small $^{1}$Department of Education Information Technology, Faculty of Education,\\
\small East China Normal University, Shanghai 200062, China\\
\small $^{2}$School of Education, Tsinghua University, Beijing 100084, China\\
\small $^{3}$University of Washington, Seattle, WA 98195, USA\\[0.4em]
\small\texttt{51274108057@stu.ecnu.edu.cn}}
\date{}
\begin{document}
\maketitle
\thispagestyle{plain}
\begin{abstract}
The transition from block-based to text-based programming requires learners to convert visible program structures into abstract textual expressions, which may create a cognitive gap between understanding computational concepts and expressing them in Python syntax. To support this transition, we designed and implemented BlockPython. The platform centers on bidirectional translation between blocks and Python and guides learners through four stages: Task Decomposition, Block-Based Practice, Code Challenge, and Extended Interaction. Across these stages, learners progressively establish connections among program structure, runtime behavior, and textual code. During learning, the platform continuously collects process evidence, including block artifacts, code versions, run outcomes, use of support, and dialogue. Deterministic diagnosis, program visualization, and the learning assistant use this evidence to identify different difficulties in computational understanding and Python expression. The rule-based system is responsible for program execution, objective evaluation, and stage control, while the learning assistant uses verified evidence to provide explanations, prompts, and guiding questions. This report describes the design rationale, learning workflow, and process-aware support mechanisms of BlockPython and provides a system-design reference for supporting the transition from block-based to text-based programming and for analyzing learning processes.
\end{abstract}
\section{Introduction}
As programming education progresses beyond introductory activities, learners are expected to move from block-based environments such as Scratch to text-based environments such as Python (Brown et al., 2022). These environments differ substantially in how program structures are represented and constructed (Weintrop \& Wilensky, 2018). Bridging the gap between the two representations and promoting effective transfer of prior knowledge therefore constitute important challenges in programming education (Strong et al., 2025).

Block-based programming represents statement order, parameter input, and structural nesting through shapes, colors, and connection mechanisms. Its operational constraints also prevent many invalid program combinations (Maloney et al., 2010). These visual cues and structural constraints can reduce the difficulty of program comprehension (Tsai et al., 2025). Python, by contrast, expresses program structure primarily through textual conventions such as variable names, parentheses, quotation marks, colons, and indentation. Learners must therefore translate structural relationships that are directly visible and manipulable in blocks into abstract textual expressions (Kölling et al., 2015). Research has shown that some learners who are familiar with basic concepts in Scratch remain unable to identify and apply the corresponding concepts in Python (Runde et al., 2023). They may also omit quotation marks or colons, confuse assignment with equality testing, or make incorrect variable assignments (Kazemitabaar et al., 2023). These patterns illustrate the cognitive gap between block-based and Python programming (Mladenović et al., 2024).

Existing approaches commonly support this transition through code conversion, side-by-side representations, or hybrid editors, yet they pay less attention to the learner's representational conversion process and to the cognitive states underlying different errors (Lin et al., 2025). BlockPython addresses this need through two complementary activities: reading Python and reconstructing the program with blocks, and studying blocks and writing the corresponding Python code. The platform combines program visualization, authentic run outcomes, and process evidence to distinguish difficulties in computational understanding from difficulties in Python expression. A deterministic rule system and a learning assistant then provide corresponding evaluation, explanation, and support.

\section{Overall System Architecture and Four-Stage Workflow}
The BlockPython student interface uses a three-pane instructional workspace (see Figure 1). The left pane contains the Blockly workspace, where students can browse the course toolbox, construct programs, or inspect read-only reference blocks. The center pane contains task instructions, target code or a code editor, run controls, immediate feedback, and program visualization. The right pane contains the learning assistant dialogue for stage explanations, error support, on-demand help, and extension-task discussion. Four stage controls remain visible at the bottom, allowing students to identify their current stage and the subsequent stages that have been unlocked.

Internally, the platform connects student artifacts, program runs, and dialogue into a continuous learning process. After students inspect decomposed task steps, the system records the concepts they have encountered. After they construct blocks, it stores stable artifacts and analyzes connections and nesting. After they run a program, it records input, output, variable changes, and the objective result. When students request help or repeat an error, the learning assistant uses this evidence to provide relevant support. At stage transitions, the back end also produces concise stage summaries for subsequent stages and teacher reports.

\begin{figure}[htbp]
\centering
\includegraphics[width=0.96\linewidth]{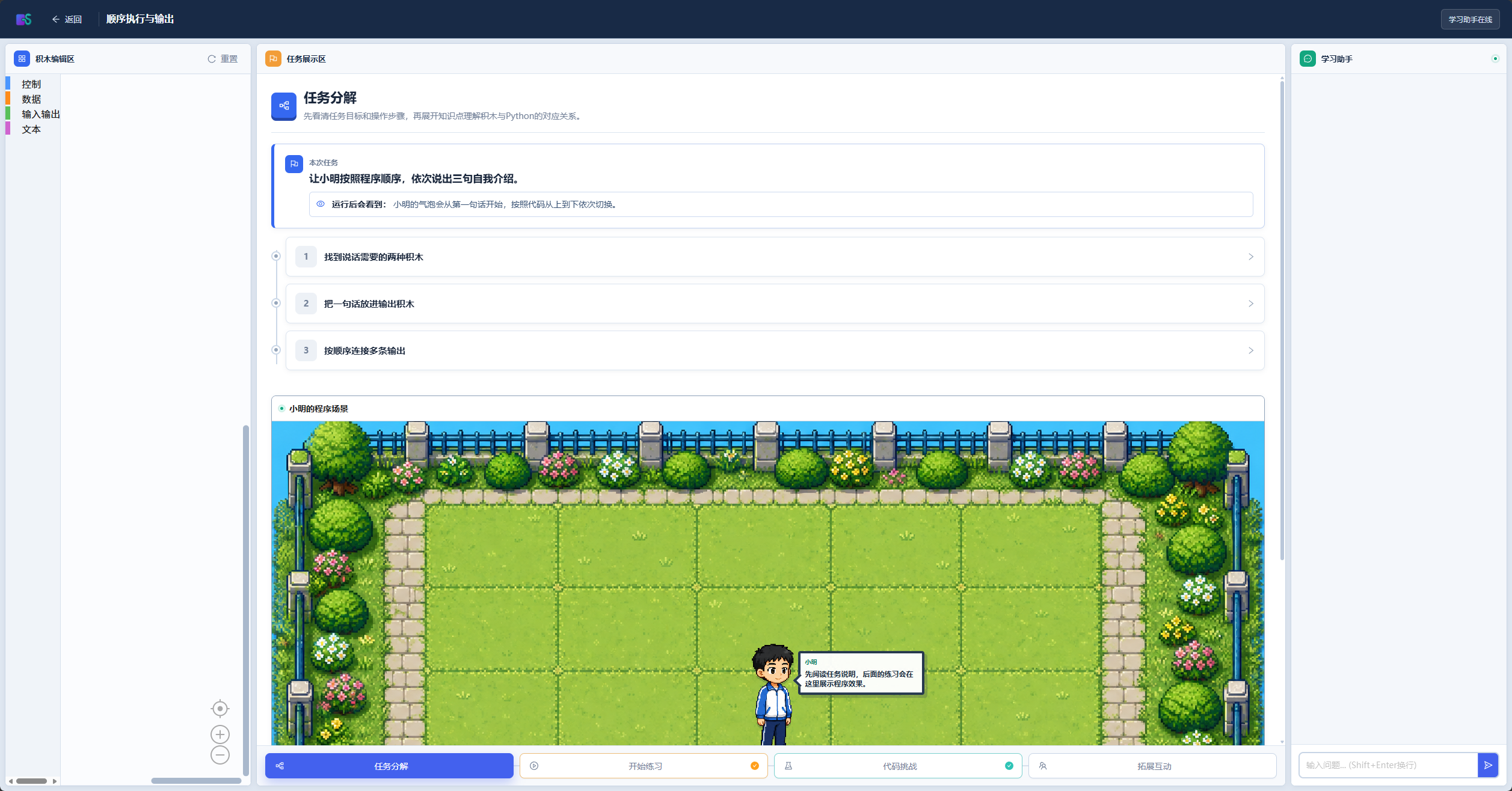}
\caption{BlockPython task interface.}
\end{figure}
\subsection{Task Decomposition Stage}
The Task Decomposition Stage helps students connect the task objective, block forms, and Python semantics. The upper part of the page states what the task should achieve and what students should observe after execution. The complete task is then divided into manageable actions, such as locating the required blocks, placing text inside an output block, and connecting statements in the intended sequence. This organization supports an understanding of local actions before students assemble the complete program structure.

Each step supports two levels of expansion (see Figure 2). The first level describes the concrete action and presents the block cards required for that step. Their colors correspond to the Control, Data, Input/Output, and Text categories in the Blockly toolbox. Selecting a block card reveals its meaning, the corresponding Python translation rule, an independent micro-example, and relevant notes on connection or indentation. Micro-examples use text and variable names that differ from the target task, reducing the likelihood that students will copy a complete answer.

The learning assistant provides explanation and orientation at this stage. After a student expands a step and views it briefly, the assistant offers a concise explanation related to the visible content and asks one observation or prediction question. For example, when a student examines an output block, the assistant explains that a text block must be placed inside it and invites the student to observe the execution order of vertically connected output blocks. Students may also ask questions in the dialogue pane.

Viewing evidence from Task Decomposition informs later support. The system distinguishes between steps the student has inspected and relevant concepts that remain unviewed. If the student later encounters a connection problem in Block-Based Practice, the learning assistant can determine whether the relationship was previously examined and decide whether a brief reminder or additional explanation is appropriate.

\begin{figure}[htbp]
\centering
\includegraphics[width=0.96\linewidth]{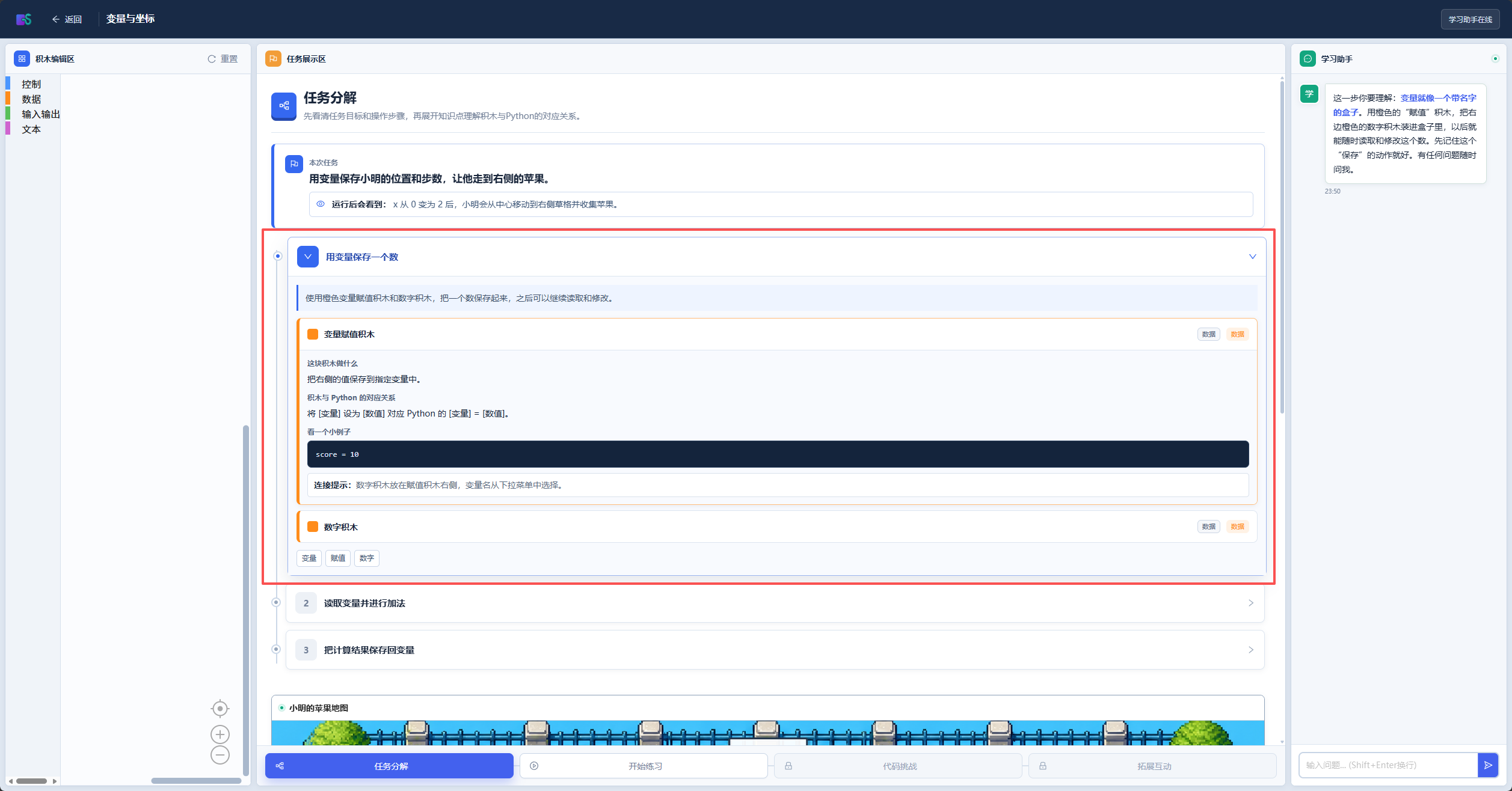}
\caption{Two-level knowledge expansion in the Task Decomposition Stage.}
\end{figure}
\subsection{Block-Based Practice Stage}
During Block-Based Practice, students read a complete target Python program and construct a semantically equivalent block program in an initially blank Blockly workspace (see Figure 3). The toolbox provides the foundational blocks used throughout the course, including variable assignment and access, numbers, arithmetic, comparisons, if--else, for--range, text, text concatenation, print, and input. Students select blocks, complete fields, connect inputs, arrange statement order, and create the required nesting.

When a student selects Generate and Run, the system first validates the block structure. It checks for disconnected statements, empty input slots, incomplete statement connections, and incorrect nesting of conditionals or loops. If the structure is valid, registered generators produce executable Python together with a mapping between code lines and source blocks. The generated code is sent to the same runtime used for student-authored Python, and the run outcome and task objective are evaluated by deterministic rules.

The first occurrence of a structural problem produces immediate deterministic feedback, such as 'The second output block is missing text' or 'The conditional block has not been placed inside the loop.' The learning assistant intervenes when the same problem recurs, when the student explicitly requests help, or when meaningful inactivity is detected. Level 1 support identifies the area or concept to inspect. Level 2 support points to a specific block, input slot, or connection. Level 3 support provides a task-independent micro-example and a stepwise checking strategy. When a verified location is available, the relevant block is selected or highlighted and a short location label appears in the workspace header.

\begin{figure}[htbp]
\centering
\includegraphics[width=0.96\linewidth]{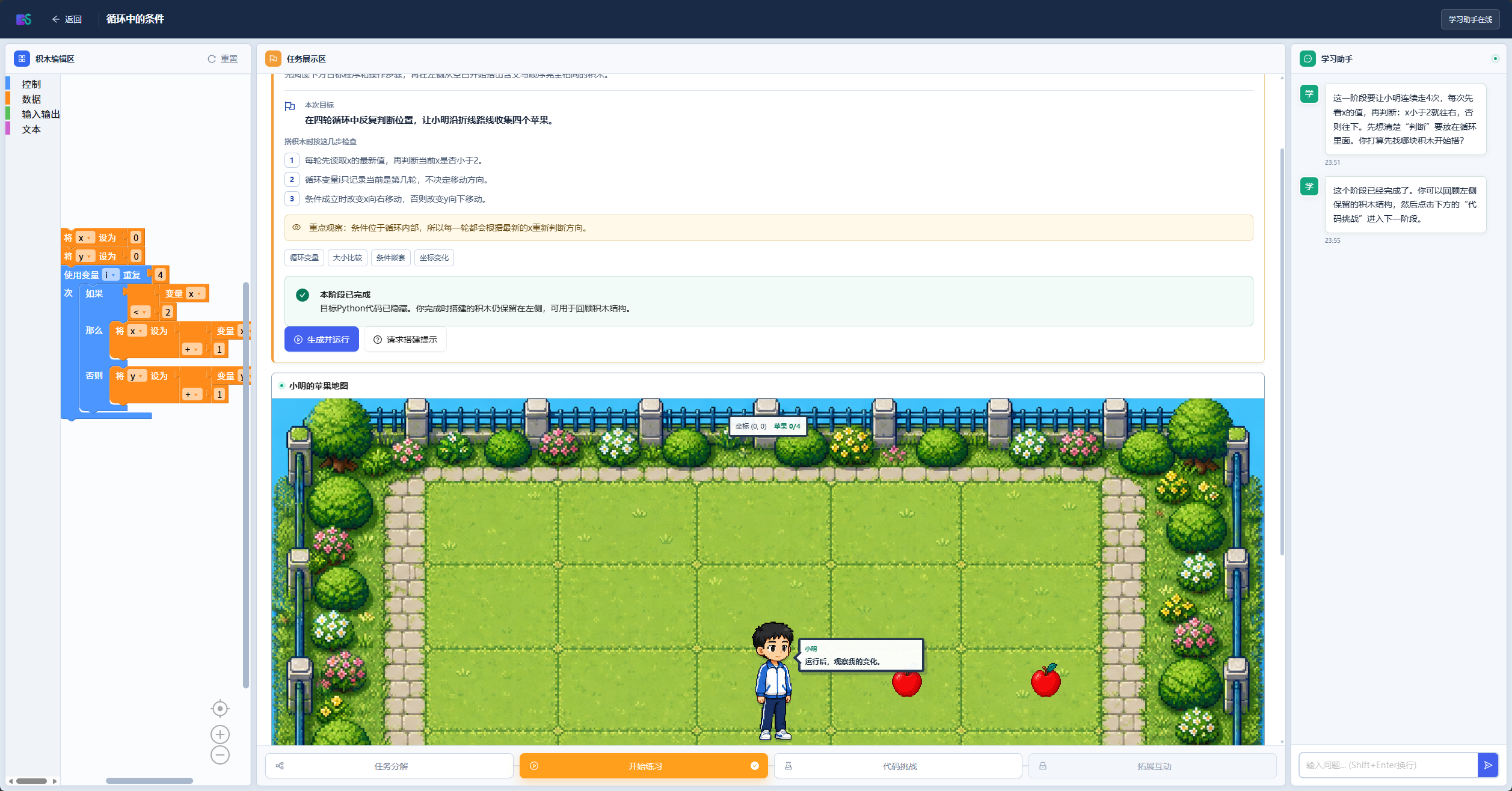}
\caption{Student artifact, diagnosis, and localized support in Block-Based Practice.}
\end{figure}
\subsection{Code Challenge Stage}
The Code Challenge Stage presents a complete, connected block program as a read-only reference. Students independently write the corresponding Python program in a blank editor. The blocks cannot be moved or edited, ensuring that all students work from the same semantic representation. Block-Based Practice and the Code Challenge use the same objective and evaluation, so students must translate statement sequence, input relationships, and nesting into valid textual code rather than relying on a change of task (see Figure 4).

When students select Run Code, the system performs syntax analysis before execution. Missing quotation marks, parentheses, colons, or indentation prevent visualization from starting; the editor identifies the actual error line and displays an understandable error category. If the code is executable, the runtime records output, variables, branch decisions, loop iterations, and final state, after which the system determines whether the program satisfies the task objective. If executable code produces an incorrect result, the authentic result is retained so that students can distinguish successful execution from successful task completion.

Support in the Code Challenge combines the final block semantics from Block-Based Practice with the structure of the current Python program. The system can identify cases in which students expressed the logic correctly with blocks but encountered a local Python syntax problem, as well as cases in which a condition, loop, or statement sequence was omitted during translation. The learning assistant therefore provides concept-oriented, transfer-oriented, or syntax-oriented support. It may identify a verified line and direct attention to punctuation or indentation without supplying the complete target program.

Support is more proactive in the Code Challenge than in Block-Based Practice because students are writing Python independently for the first time. After an unsuccessful run, the system provides clear line-level deterministic feedback. A repeated diagnosis triggers a more specific Level 2 prompt, followed by a Level 3 micro-example if the difficulty persists. If nonempty code contains an unresolved problem and remains unchanged for approximately 20 seconds, the system may issue one inactivity prompt. Prompts are suppressed while the student is typing, running the program, viewing the animation, or communicating with the assistant.

\begin{figure}[htbp]
\centering
\includegraphics[width=0.96\linewidth]{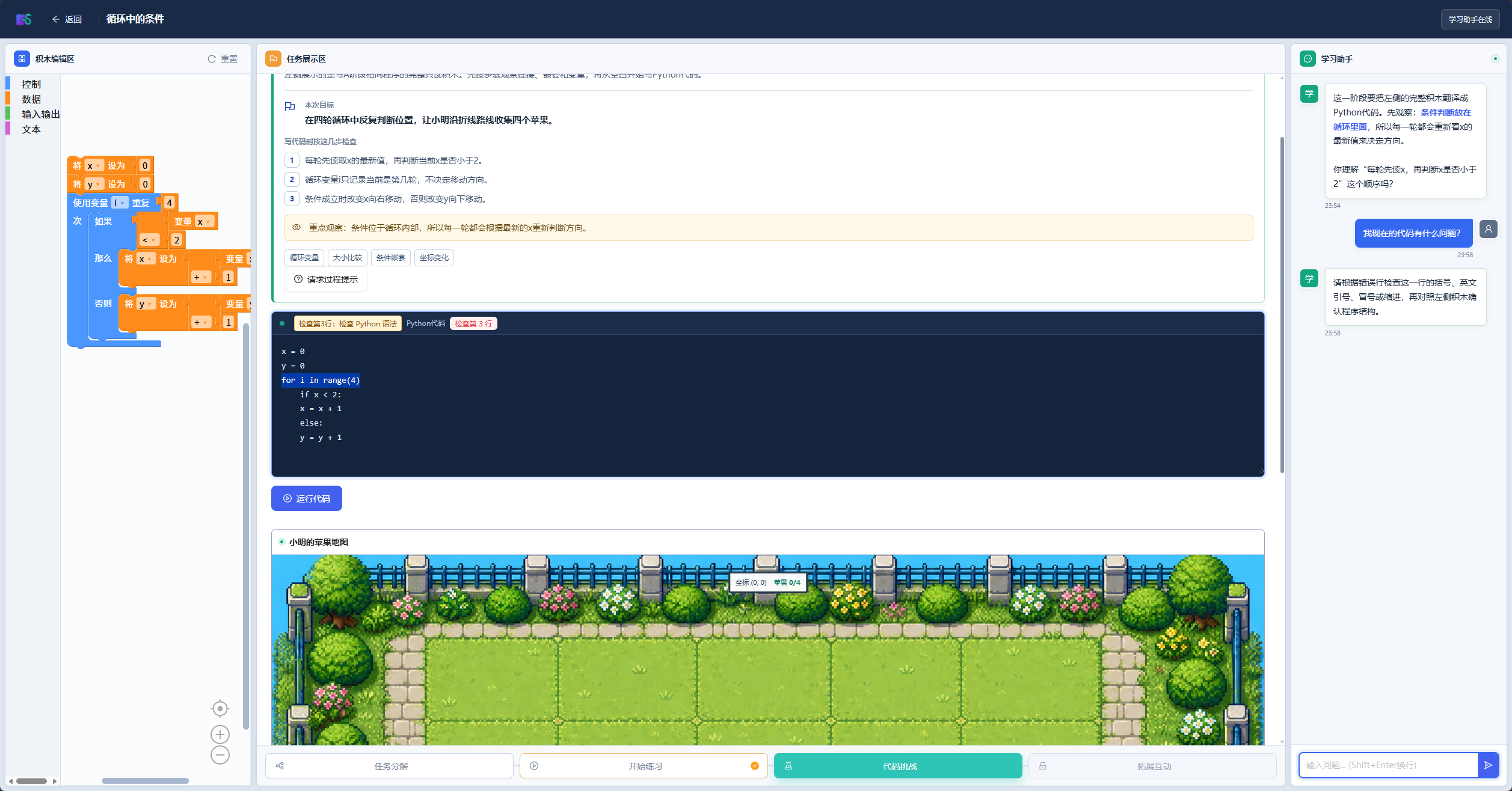}
\caption{Read-only blocks and line-level support in the Code Challenge Stage.}
\end{figure}
\subsection{Extended Interaction Stage}
The Extended Interaction Stage adds a bounded requirement to the original task, such as an additional output statement, a revised route order, an extra variable, or another fruit to collect. All students receive the same extension objective and success criteria. The system adapts only the focus of discussion, question framing, support level, and intervention timing. This preserves experimental comparability while enabling evidence-informed personalization (see Figure 5).

Students begin by selecting Discuss My Approach with the Learning Assistant. The assistant reads valid summaries from the preceding three stages and the cross-task learning profile, selects one primary support focus, and asks one concise question. Dialogue follows a coherent sequence: reviewing authentic prior work, eliciting the student's explanation, asking a targeted follow-up question, supporting independent implementation, examining a program run, reflecting jointly, and summarizing the exchange. Students write code in a separate Python editor. After each run, the assistant reads only the latest result that matches the current artifact. Following success, it asks the student to explain why the key structure worked. Following failure, it helps the student compare the intended behavior with the actual output, variable trace, or route before another revision. Dialogue state is stored by student and task, allowing the interaction to continue coherently after a page refresh or later return.

The 60-second inactivity mechanism operates only when the student has a nonempty artifact, an unfinished interaction goal, and a visible page. It is suppressed while the student is editing, viewing an animation, entering program input, or composing a message. The assistant does not recite a chronological list of prior runs and does not expose internal learning profiles or knowledge-state labels to students.

\begin{figure}[htbp]
\centering
\includegraphics[width=0.96\linewidth]{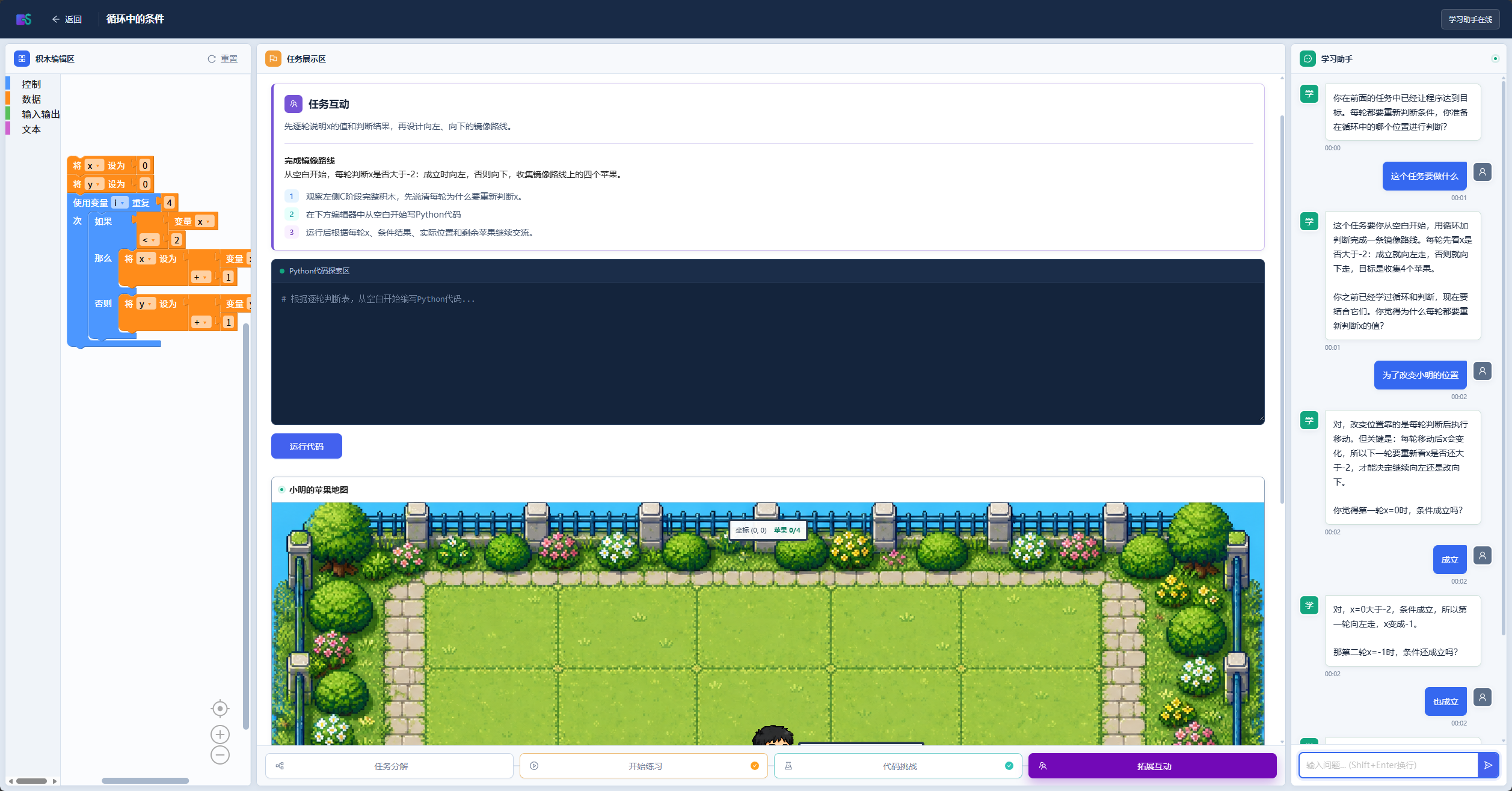}
\caption{Personalized, continuous dialogue in the Extended Interaction Stage.}
\end{figure}
\section{Six Tasks and the Program Visualization Mechanism}
The six tasks follow prerequisite relationships among programming concepts. Task 1 uses three print statements to introduce output and sequential execution. Task 2 assigns x, y, and step variables and changes a position through arithmetic. Task 3 uses input to receive a fruit name and if--else to select a route. Task 4 uses for--range to repeat a round trip. Task 5 places a conditional inside a loop and evaluates the current variable value on each iteration. Task 6 integrates input, conditionals, loops, and two-dimensional coordinates in a route-planning task involving two pieces of fruit. Tasks 3 and 6 are designed for two lessons each, while the other tasks require one lesson each, producing an eight-lesson sequence.

The tasks share stable visual conventions. In coordinate-based tasks, Xiaoming begins on the central tile at (0, 0). Increasing x moves the character to the right, decreasing x moves the character to the left, increasing y moves the character downward, and decreasing y moves the character upward. One unit corresponds to one grass tile. Output from print appears in a speech bubble beside Xiaoming. When the character reaches a fruit tile, the interface displays the collection event. Task 1 hides coordinates and fruit and presents only speech bubbles and an execution-order summary (see Figure 6).

Visualization is driven by authentic runtime events. The runtime records the active Python line, variable values before and after each change, output, branch decisions, and loop events. A local front-end player converts changes in x and y into continuous movement, print events into speech bubbles, and entry into a fruit tile into a collection animation. When a value change spans multiple tiles, the character pauses for one second on each tile so that students can observe the transition.

The platform distinguishes among code that cannot execute, a runtime failure, and behavior that does not meet the task objective. Syntax errors do not trigger animation because the textual form must first be corrected. A runtime error preserves the events completed before failure. Executable code with incorrect output or an incorrect route displays the authentic behavior and reports an objective mismatch after playback. At the end of the animation, the interface retains the final coordinates, key variable change, collection state, and task outcome for further analysis.

Task completion is based on behavioral objectives. Deterministic evaluators inspect actual output, final variables, route behavior, collection results, required structures, and multiple test cases for input-based tasks. A student may succeed with an allowed implementation that differs from the reference solution.

\begin{figure}[htbp]
\centering
\includegraphics[width=0.96\linewidth]{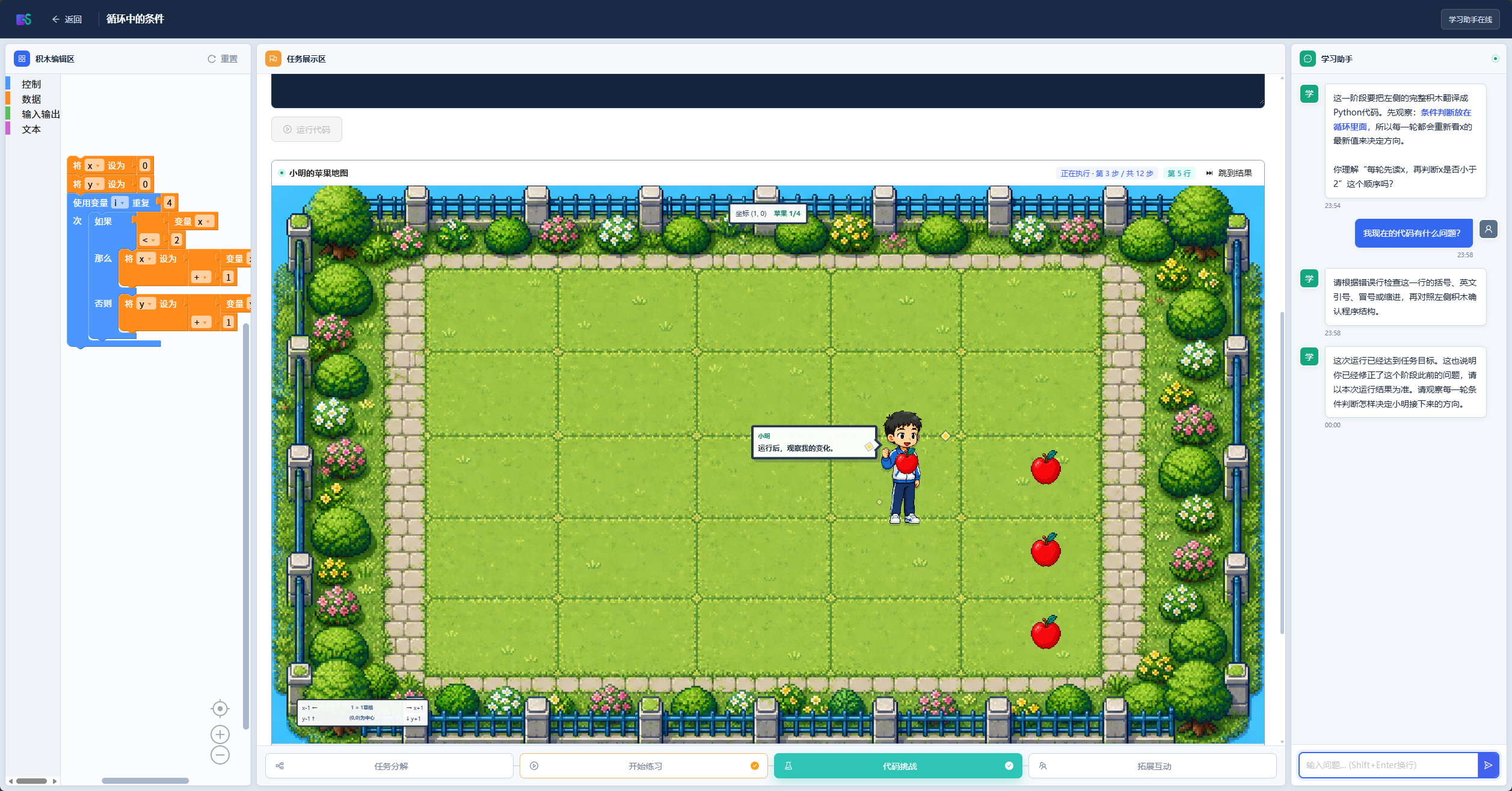}
\caption{Program visualization driven by authentic execution traces.}
\end{figure}
\section{Process-Aware Learning Assistant Mechanism}
Learning support is divided across three layers. The deterministic rule system analyzes artifacts, executes programs, evaluates objectives, diagnoses errors, enforces stage permissions, and selects support levels. A back-end summarization agent compresses stage evidence into stage summaries, task-level learning profiles, cross-task learning profiles, and teacher-facing instructional recommendations. The student-facing learning assistant uses the current artifact, matching run result, deterministic diagnosis, and valid summaries to explain concepts and conduct dialogue in language suitable for lower-secondary students.

This division maintains a stable source of truth. A language model cannot classify an incorrect program as correct, unlock a stage, or modify a student artifact. Statements about block count, text values, variable values, code lines, error locations, and task semantics are checked against the current artifact. A response that fails validation is retried once; if it remains unreliable, the system returns deterministic process guidance. If the summarization service times out or produces invalid content, the back end creates a template-based summary while student execution and artifact storage continue normally.

Collected process data include the task steps and concepts viewed during Task Decomposition; block artifacts, connections, inputs, nesting, and generated code from Block-Based Practice; code versions, syntax structures, and meaningful edits from the Code Challenge and Extended Interaction; run inputs, outputs, errors, variables, branches, loops, and final states; support requests, support levels, interface locations, and subsequent revisions; stage-specific dialogue; and active learning time.

A stable artifact snapshot is created approximately 1.5 seconds after editing stops and is saved immediately upon submission, execution, help requests, stage changes, and task exit. Identical content is deduplicated. Every run and agent request is bound to the task, stage, and artifact version. When the artifact changes, obsolete run results, prompts, and highlights cannot be used to describe the current work. Front-end action locks and back-end idempotency merge repeated rapid clicks, and the interface directly asks students to avoid repeated clicking instead of generating multiple model responses.

A stage summary typically uses four or five sentences to describe the concepts viewed, artifact structure, principal difficulty, use of support, revision behavior, and a next-step recommendation. The task-level learning profile combines evidence from the four stages of one task. The cross-task learning profile records recurring difficulties, independent successes, and changes in reliance on support across multiple tasks. Every profile conclusion is associated with an evidence version and can be traced to a block artifact, code version, run result, or dialogue exchange.

Learning profiles determine support priorities. For example, two students may receive the same Task 5 extension. If one has repeatedly struggled to read changing variable values, the assistant prioritizes an explanation of the value on each iteration. If the other built the block structure correctly but repeatedly made indentation errors in Python, the assistant prioritizes comparison between block nesting and textual indentation. Both students retain the same objective and success criteria.

The effect of a prompt is inferred from subsequent changes to the artifact. If the student modifies the indicated structure and resolves the diagnosis, the prompt is recorded as adopted. If the relevant structure changes but the problem remains, it is partially adopted. If several meaningful edits are unrelated to the indicated area, the prompt is treated as ignored. If no subsequent activity is observed, the outcome remains unobserved. These records help reduce repetitive support, update learning evidence, and help teachers understand students' use of scaffolding.

\section{Learning-State Persistence and Stage Permissions}
The platform stores task-specific learning state for each student. This includes completion status and timestamps for Block-Based Practice and the Code Challenge, whether the target Python reference has been hidden, the latest block artifact, Code Challenge code, Extended Interaction code, the most recent update time, and stage-specific dialogue. Drafts are saved after stable editing and immediately when students submit, run, change stage, or leave a task. Artifacts, permissions, and dialogue can therefore be restored after logout or a later return.

Whenever students enter a task from the task center, the Task Decomposition Stage is shown first so that the task objective is visible again. Temporary interface state, including scroll position, cursor position, expanded cards, and animation progress, is not restored. Completed work and artifacts from Block-Based Practice and the Code Challenge remain available, and students may continue to enter stages that have already been unlocked.

Task Decomposition and Block-Based Practice are initially accessible. Successful completion of Block-Based Practice unlocks the Code Challenge, and successful completion of the Code Challenge unlocks Extended Interaction. Locked controls explain which preceding stage must be completed, and the back-end run and dialogue endpoints enforce the same permissions. After a student first enters the Code Challenge, the target Python program in Block-Based Practice is replaced with a completion message, reducing the opportunity to copy the reference program. The student's final blocks remain available as a read-only artifact.

Reset behavior is tailored to each stage. Before Block-Based Practice is completed, Reset clears the editable workspace. After completion, the blocks become read-only and Reset is disabled. Reference blocks in Task Decomposition and the Code Challenge remain read-only. In Extended Interaction, Reset restores and recenters the complete Code Challenge blocks while preserving the student's extension code.

\section{Teacher-Facing Learning Process Reports}
The teacher interface organizes learning evidence by student and by task. Each task report contains three areas: Task Overview, Artifacts and Process, and Support and Dialogue (see Figure 7). Task Overview presents active learning time, stage-level time, completion of Block-Based Practice and the Code Challenge, number of attempts, help requests, principal difficulties, and a concise learning-evidence summary. This provides a rapid account of the student's overall work on the task.

Artifacts and Process presents consolidated key attempts in chronological order. Block-Based Practice artifacts are reconstructed in a read-only Blockly canvas with relevant problem locations highlighted (see Figure 8). Code Challenge and Extended Interaction artifacts are presented as formatted code together with program input, output, errors, and run outcomes. Internal locations are translated into teacher-readable descriptions such as 'the third output block,' 'inside the conditional block,' or 'Python line 4.' Identical snapshots, repetitive edits, and low-value system events are consolidated automatically.

Support and Dialogue displays conversation bubbles grouped by task and stage. Teachers can see when the assistant intervened, which verified difficulty triggered the intervention, what support level was used, and whether the student subsequently revised the relevant structure. Stage summaries and the task-level learning profile are displayed alongside links to the corresponding artifact or run evidence.

Teachers may request a concise instructional recommendation. The back-end summarization agent organizes deterministic evidence into observed strengths, current difficulties, block-to-Python transfer, reliance on support, and recommended next instructional steps.

\begin{figure}[htbp]
\centering
\includegraphics[width=0.96\linewidth]{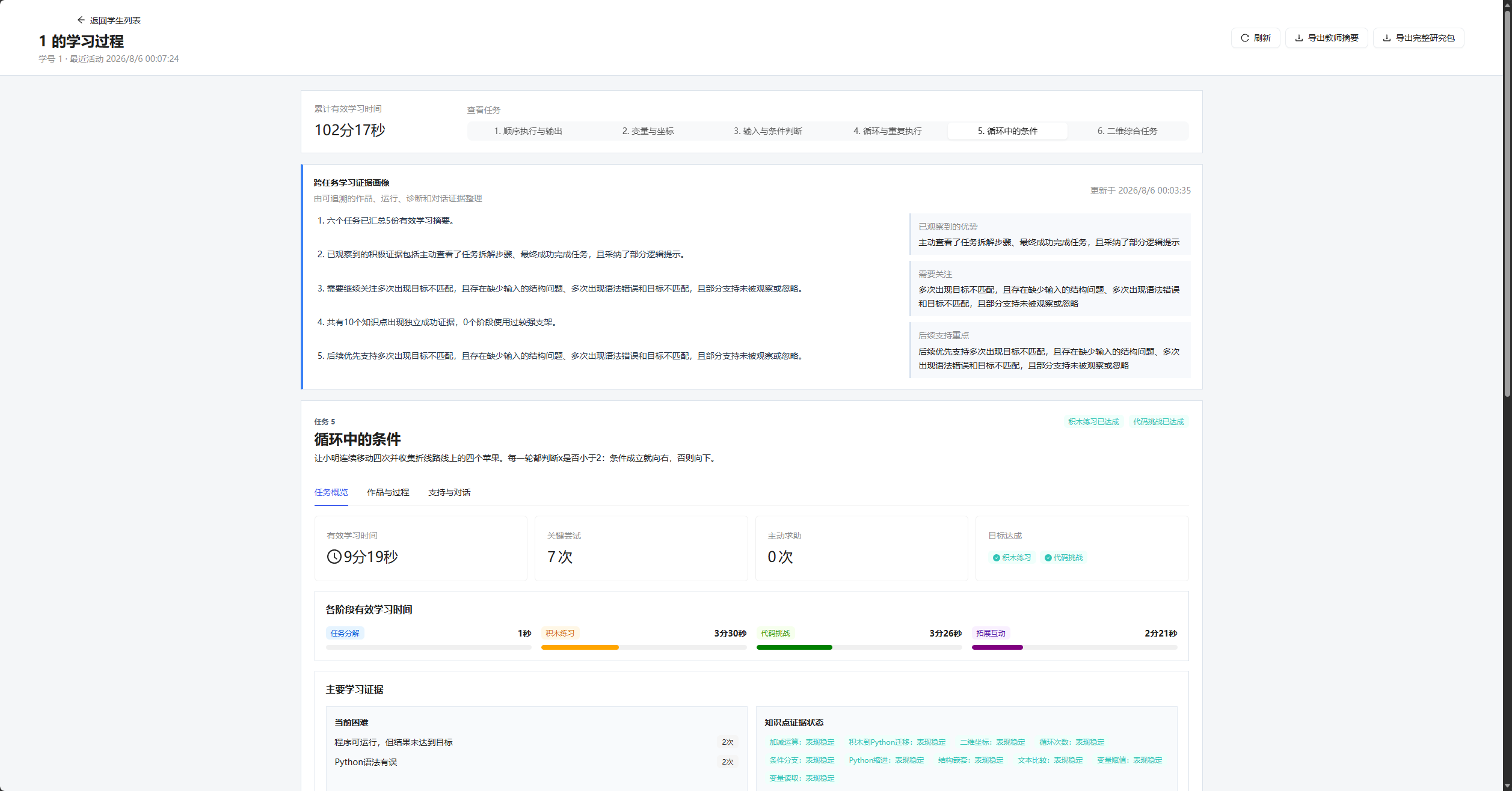}
\caption{Teacher-facing report for a single student task.}
\end{figure}
\begin{figure}[htbp]
\centering
\includegraphics[width=0.96\linewidth]{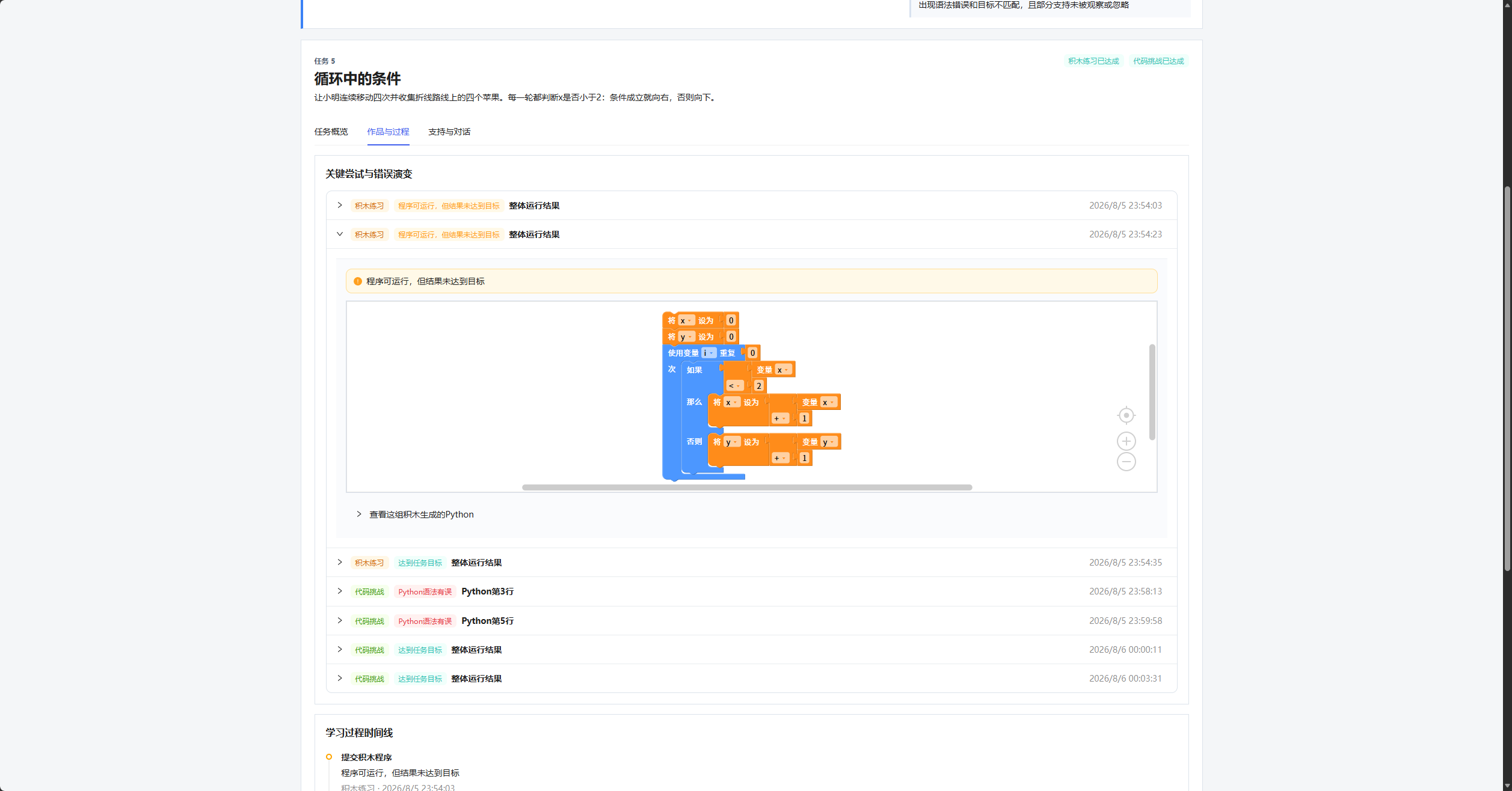}
\caption{Student artifacts and support process in the teacher interface.}
\end{figure}
The teacher interface provides two export formats. CSV contains a readable task-level summary suitable for routine instructional use. A complete JSON research package retains raw events, artifact versions, XML, hashes, and model audit information for research analysis and system review. Separating routine reporting from research export keeps the teacher interface understandable while preserving traceability.

\section{System Implementation}
The student front end is implemented with React, TypeScript, Vite, Ant Design, Blockly, Zustand, and Axios. The back end uses Node.js, Express, TypeScript, and MySQL and executes student programs in separate Python processes. The three-pane instructional interface is designed for common classroom desktop resolutions. The login page, task center, instructional workspace, and teacher pages use route-based loading, while Blockly and teacher-facing dependencies are bundled separately.

The student-facing learning assistant and back-end summarization agent use independent model configurations. The student assistant supports real-time interaction within the current task, whereas the summarization agent produces stage summaries, learning profiles, and teacher recommendations. Failure of either model service does not block block-to-Python generation, Python execution, deterministic evaluation, stage unlocking, or artifact storage. The health endpoint reports course and agent-policy versions to support deployment checks.

The controlled runtime supports only the Python constructs required by the six tasks and applies a five-second execution limit and a one-megabyte output limit. A run is terminated when the page is left, the request is interrupted, or the time limit is exceeded. Input-based tasks open an in-system prompt and send the student's response to Python as standard input. Runtime events are buffered on the front end and controlled by a local player, so pause, replay, and Skip to Result do not repeat server-side execution.

\end{document}